\documentclass[10pt,twocolumn,letterpaper]{article}

\usepackage[pagenumbers]{wacv} 

\usepackage{graphicx}
\usepackage{amsmath}
\usepackage{amssymb}
\usepackage{booktabs}
\usepackage{algorithm}
\usepackage[noend]{algorithmic}

\usepackage[pagebackref,breaklinks,colorlinks]{hyperref}

\usepackage[capitalize]{cleveref}
\crefname{section}{Sec.}{Secs.}
\Crefname{section}{Section}{Sections}
\Crefname{table}{Table}{Tables}
\crefname{table}{Tab.}{Tabs.}

\def\wacvPaperID{851} 
\def\confName{WACV}
\def\confYear{2024}

\begin{document}

\title{Unsupervised 3D Pose Estimation with Non-Rigid Structure-from-Motion Modeling}

\author{Haorui Ji$^{1}$ \quad Hui Deng$^{2}$ \quad Yuchao Dai$^{2}$ \quad Hongdong Li$^{1}$  \\
$^{1}$The Australian National University\\
$^{2}$School of Electronics and Information, Northwestern Polytechnical University \\
{\tt\small \{haorui.ji, hongdong.li\}@anu.edu.au, denghui986@foxmail.com, daiyuchao@gmail.com}
}
\maketitle

\begin{abstract}
   Most of the previous 3D human pose estimation work relied on the powerful memory capability of the network to obtain suitable 2D-3D mappings from the training data. Few works have studied the modeling of human posture deformation in motion. In this paper, we propose a new modeling method for human pose deformations and design an accompanying diffusion-based motion prior. Inspired by the field of non-rigid structure-from-motion, we divide the task of reconstructing 3D human skeletons in motion into the estimation of a 3D reference skeleton, and a frame-by-frame skeleton deformation.  A mixed spatial-temporal \textbf{NRSfMformer} is used to simultaneously estimate the 3D reference skeleton and the skeleton deformation of each frame from 2D observations sequence, and then sum them to obtain the pose of each frame. Subsequently, a loss term based on the diffusion model is used to ensure that the pipeline learns the correct prior motion knowledge. Finally, we have evaluated our proposed method on mainstream datasets and obtained superior results outperforming the state-of-the-art.
\end{abstract}

\section{Introduction}
\label{sec:intro}

\begin{figure}[t]
 \centering
 \begin{subfigure}[t]{0.45\linewidth}
  \centering
  \includegraphics[width=\textwidth]{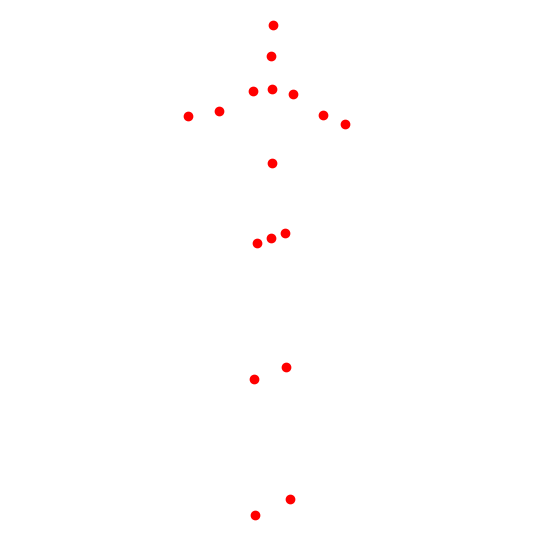}
  \caption{2D Image Keypoints}
 \end{subfigure}
 \hfill
 \begin{subfigure}[t]{0.45\linewidth}
  \centering
  \includegraphics[width=\textwidth]{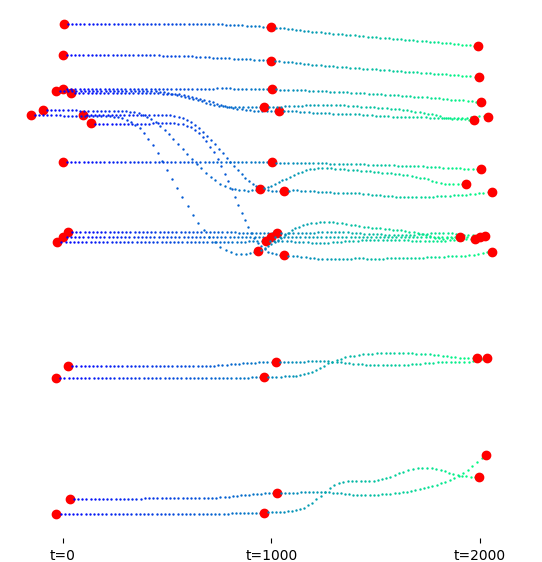}
  \caption{2D Keypoints Tracking Across Frames}
 \end{subfigure}
  \begin{subfigure}[t]{0.45\linewidth}
  \centering
  \includegraphics[width=\textwidth]{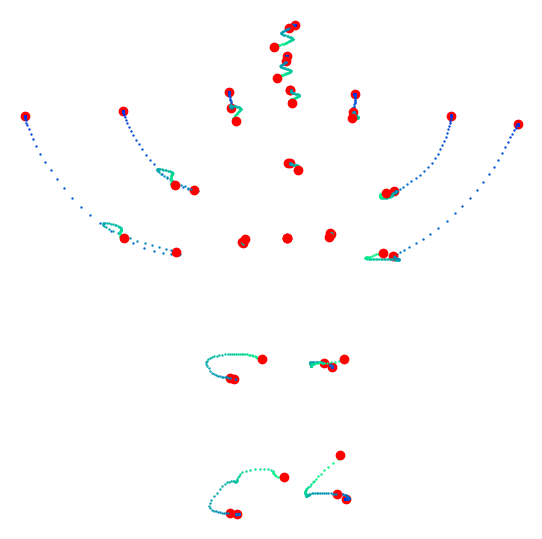}
  \caption{Centralized 2D Keypoints Tracks}
 \end{subfigure}
 \hfill
  \begin{subfigure}[t]{0.45\linewidth}
  \centering
  \includegraphics[width=\textwidth]{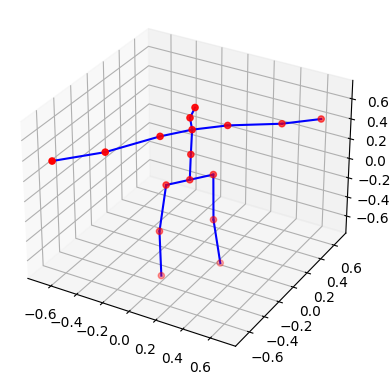}
  \caption{3D Reconstruction of Non-Rigid Moving Object}
 \end{subfigure}
 \caption{A visual illustration of basic pipeline setup for solving 3D HPE problem with NRSfM factorization. Given a sequence of 2D keypoints to form a track that describes the motion of a non-rigidly deforming subject, NRSfM aims to reconstruct the structure and pose of this subject in 3D space at each timestamp.}
 \label{fig:nrsfm vis}
\vspace{-1em}
\end{figure}

Multi-frame 3D Human pose estimation (HPE) aims to recover human 3D poses from 2D image sequences. It has a wide range of applications such as augmented reality, robotics, sports analysis and film industry, \etc. \cite{wang2020deep, kocabas2020vibe, golyanik2016nrsfm, jiang2022golfpose}.   3D HPE can be viewed as a special instance of NRSfM (non-rigid structure from motion) and thus its solution may be benefited from the advances in the field of structure from motion . Fig.\ref{fig:nrsfm vis} demonstrates a typical pipeline for solving the 3D HPE problem with NRSfM.  In this paper, we will use the term "pose", "skeleton" or human "shape" interchangeably without loss of generality.

Non-rigid structure from motion is significantly more challenging than its rigid counterpart \cite{ozyecsil2017survey} because the human body can go through arbitrary rigid movements and non-rigid articulations across different viewpoints, therefore introducing multiple ambiguities: (i) depth ambiguity: single 2D measurement can correspond to multiple 3D poses and the variance across viewpoints cannot be modeled by mere multiview geometry. (ii) motion ambiguity: relative rigid motion between the human body and the camera is entangled so that absolute human pose and motion decomposition cannot be achieved. These ambiguities make NRSfM essentially an underconstrained problem and additional constraints are required to make it solvable and computationally tractable.

Accordingly, NRSfM methods focus on imposing priors in shapes and trajectories to ensure the uniqueness and correctness of the solution. On one hand, traditional optimization-based methods use hand-crafted constraints to reveal geometric properties of the deforming shapes to better regularize the optimization process. On the other hand, in recent years learning-based NRSfM methods gradually boost the performance. They focus more on using the expressiveness of neural networks to directly learn a reasonable shape and motion prior from massive amounts of data.

In this paper, we aim to address the 3D HPE problem in an NRSfM manner. Many existing 3D HPE methods rely heavily on the powerful memory capability of the neural networks to obtain suitable 2D-3D lifting mappings from the training data, few have studied the modeling of human posture deformation in motion. Inspired by the NRSfM problem, we design a geometric-aware 3D HPE pipeline, called RMNRD (standing for: Rigid-Motion-Non-Rigid-Deformation), that divides the human pose sequence lifting into the estimation of a 3D sequential e reference skeleton, and the frame-by-frame skeleton deformation. In addition, to alleviate the inherent depth ambiguity, we propose to utilize a diffusion-based pose generator and leverage the prior knowledge encoded within to enforce our network to predict reasonable 3D poses conditioned on 2D observations but without any 3D supervision. In summary, our contributions are as follows:
\begin{itemize}
    \item We propose RMNRD, an NRSfM-style strategy to model the 3D human pose estimation by dividing the sequence lifting problem into the estimation of sequential reference skeleton, skeleton deformation of each frame, and corresponding rigid transformation.  
    \item We propose NRSfMFormer, a mixed spatial-temporal architecture that implements the RMNRD strategy, estimating the reference skeleton, skeleton deformation, and rigid transformation simultaneously.
    \item We utilize diffusion models trained on large human pose generation datasets to embed motion knowledge prior. It enforces our network to predict 3D poses that are consistent with the prior but without any 3D annotations.
    \item Our method achieves comparably good performance across multiple benchmarks compared with State-Of-The-Art(SOTA) deep NRSfM methods, indicating the effectiveness of our method in dealing with sequential data.
\end{itemize}

\section{Related Work}
\label{related work}
\noindent\textbf{Unsupervised 3D HPE} In contrast to supervised 3D HPE, unsupervised setting does not allow the usage of 3D annotation, which makes the problem much more challenging due to the lack of data to learn the one-on-one 2D-3D mapping. Recent methods utilize multi-view information as a supervision signal and try to alleviate the depth ambiguity problem. Rhodin \etal\cite{rhodin2018unsupervised} firstly propose to learn a geometry-aware skeleton structure constructed via multiview images. Other methods \cite{chen2019unsupervised, wandt2022elepose} bypass the reliance on multi-view inputs and construct self-consistency in 3D space in a loop-closure manner with random rotation perturbation. Other priors have also been considered such as adversarial training\cite{chen2019unsupervised}, kinematic human skeleton structure\cite{kundu2020kinematic}, or fixed scale bone-length ratios\cite{yu2021towards} to enforce the predicted human pose to be realistic and accurate. 

\noindent\textbf{Non-Rigid Struture-from-Motion} Recovering 3D shape and motion from 2D keypoint sequences has a long research history. Many works have constructed different sequence modeling to solve this problem. For example, deforming shapes should span a low-rank space\cite{bregler2000recovering, dai2014simple, akhter2008nonrigid}, deform smoothly\cite{gotardo2011computing}, lie in a union of linear subspaces\cite{zhu2014complex}, or satisfy certain probability distribution\cite{lee2013procrustean, lee2014procrustean}, etc. Other priors such as modeling object trajectories with DCT bases\cite{akhter2008nonrigid} and Dai \etal.'s block-matrix rank-minimization \cite{dai2014simple} have also been explored. These optimization-based methods are sensitive to initialization, and the calculation of optimization-based methods is time-consuming.
Recently, researchers resort to neural networks to learn shape and motion priors directly from large-scale data. Novotny \etal.\cite{novotny2019c3dpo} and follow-up approaches \cite{zeng2022mhr, zeng2021pr, wang2021paul, deng2022deep} introduced the transversal property to constrain shapes to a canonical view. In contrast, Park \etal.\cite{park2020procrustean} targeted NRSfM through shape alignment, constraining motion parameters so that sequential 3D structures will align to a reference frame. Furthermore, \cite{kong2020deep, wang2020deep} successfully incorporate perspective projection in learning-based NRSfM framework and is able to handle missing data. These two sets of methods inspire our Rigid-Motion-Non-Rigid-Deformation (RMNRD) modeling strategy, which takes entire sequences as input for more informed canonical view generation.

\noindent\textbf{Diffusion-based generative models} Diffusion models are a family of generative models that are inspired by non-equilibrium thermodynamics\cite{ho2020denoising, song2020denoising}. They define a Markov chain of diffusion steps to slowly add random noise to data and then learn to reverse the diffusion process to construct desired data samples from the noise. For their simple designs and flexibility in describing the underlying structure in arbitrary data, diffusion models have sparked remarkable progress in numerous areas, including image generation\cite{rombach2022high, nichol2021glide, ramesh2022hierarchical, saharia2022photorealistic}, object detection\cite{chen2022diffusiondet}, semantic segmentation\cite{brempong2022denoising, xu2023open}, etc. In addition to the success in the 2D image domain, more recent works have extended diffusion models to 3D content generation with multi-modality inputs. For instance, \cite{shan2023diffusion, ci2023gfpose} apply diffusion models to 3D human pose estimation to resolve depth ambiguity, and \cite{tevet2022human, shafir2023human, zhang2022motiondiffuse} use them as priors to guide 3D motion generation. In this paper, we use the diffusion model to define a plausible distribution in 3D and 2D joint locations, resulting in a novel method for lifting.

\section{Method}

\begin{figure*}[t]
\begin{center}
\includegraphics[width=\linewidth]{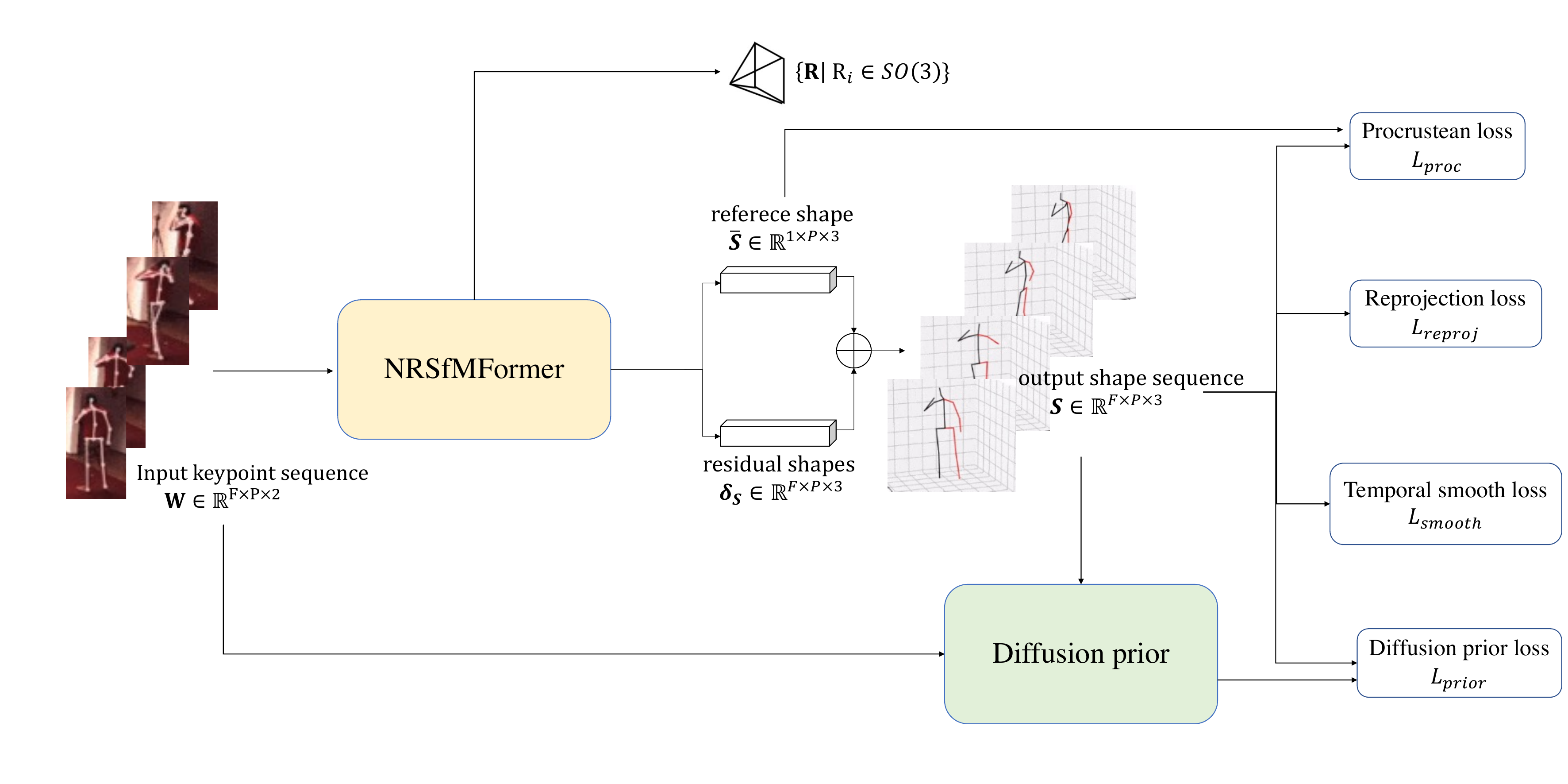}
\caption{Overview of our framework and training pipeline. Two core components of the framework are the NRSfMFormer, a transformer-based context modeling network, and a pretrained 3D pose generation diffusion model.  We apply the typical factorization paradigm but only to the sequence input and regularize it with additional constraints introduced in the loss functions in the figure. The diffusion model serves as a prior and is only used during training. In inference time, we only use the NRSfMFormer part.}
\label{fig:Pipeline}
\end{center}
\vspace{-1em}
\end{figure*}

The overall pipeline is illustrated in Fig.~\ref{fig:Pipeline}. In the following sections, we'll get down to each individual building block. Specifically, we'll introduce the novel RMNRD strategy for modeling the human pose deformation in motion in Sec.\ref{disentagle}. Next, we'll clarify how to utilize diffusion priors to resolve shape ambiguities in Sec.\ref{diffusion}. In addition, we'll address our overall network design and geometric constraints used to facilitate the optimization process in Sec.\ref{NRSfMFormer} and \ref{Loss}.

\subsection{Problem Statement}
\label{problem formulation}
We first review the NRSfM-styled 3D HPE problem formulation that will be used later to describe our approach. Consider $P$ landmarks on human skeleton tracked across $F$ frames. Observation of these landmarks from each viewpoint is represented by a measurement matrix $\mathbf{W} \in \mathbb{R}^{2F \times P}$ where each element $\mathbf{W}_i \in \mathbb{R}^{2 \times P}, i=1,\dots,F$ is generated through a camera projection $\Pi$ from the 3D point set $\mathbf{S}_i \in \mathbb{R}^{3 \times P}$, and corresponding rigid motion $(\mathbf{R}_i,\mathbf{T}_i) \in \text{SO}(3) \times \text{T}(3)$. Based on the linear model\cite{bregler2000recovering}, each non-rigid shape $\mathbf{S}_i$ can be described as a linear combination of a basis set $\{\mathbf{B}_{i1}, \mathbf{B}_{i2}, \dots \mathbf{B}_{iK}\}$ with weight coefficients $\{c_{i1}, c_{i2}, \dots c_{iK}\}$.
\begin{equation}
\label{eqn:linear combination}
    \mathbf{S}_i = \sum_{k=1}^K c_{ik}\mathbf{B}_{ik}, c_{ik} \in \mathbb{R}, \mathbf{B}_{ik} \in \mathbb{R}^{3 \times P}
\end{equation}
Assuming the measurement matrix and shape structure are already centralized, rigid motion will reduce to pure rotation, eliminating the impact of translation term. If further consider an orthographic camera model, the camera projection will become a linear function, yielding the following formulation
\begin{equation}
\begin{aligned}
\label{eqn:nrsfm}
    \mathbf{W} &=
    \begin{bmatrix} 
        x_{11}&\cdots&x_{1P}\\
        \vdots& & \vdots\\ 
        x_{F1}&\cdots&x_{FP}
    \end{bmatrix} 
    =  
    \begin{bmatrix} \mathbf{R}_1\mathbf{S}_1\\ \vdots\\ \mathbf{R}_F\mathbf{S}j_F \end{bmatrix} \\
    &=
    \begin{bmatrix} 
        c_{11}\mathbf{R}_1&\cdots&c_{1K}\mathbf{R}_1\\
        \vdots& & \vdots\\ 
        c_{F1}\mathbf{R}_F&\cdots&c_{FK}\mathbf{R}_F
    \end{bmatrix}
    \begin{bmatrix} \mathbf{B}_1\\ \vdots\\ \mathbf{B}_K \end{bmatrix} \\
    &=
    \mathbf{R}(\mathbf{C} \otimes \mathbf{I}_3)\mathbf{B}
    = \mathbf{MB}
\end{aligned}
\end{equation}
In Eqn.\ref{eqn:nrsfm}, $\mathbf{R} = \text{blkdiag}(\mathbf{R}_1,\dots,\mathbf{R}_F) \in \mathbb{R}^{2F \times 3F}$ is the camera motion matrix, $\mathbf{B} \in \mathbb{R}^{3F \times P}$ and $\mathbf{C} \in \mathbb{R}^{F \times K}$ are the matrix form of base shapes and corresponding weights coefficients. The goal of NRSfM task is to recover both motion and shape matrices $\mathbf{M}, \mathbf{B}$ from the measurement matrix $\mathbf{W}$.

\subsection{Disentangle Rigid Motion and Non-Rigid Deformation}
\label{disentagle}
One of the biggest challenges in NRSfM is the non-uniqueness in the factorization of the measurement matrix $\mathbf{W=RS}$ as in Eqn.\ref{eqn:nrsfm}. 
Intuitively, we can interpret this ambiguity in the sense that when both the human body and camera are in motion, it's impossible to discern whether the observed rigid motion originates from the human or the camera solely through 2D measurements as $\mathbf{W=RS=R'R_{camera}R_{human}S'}$. This inherent ambiguity largely accounts for why non-rigid factorization is substantially more challenging than its rigid counterpart.

In this paper, we propose a way to alleviate this ambiguity by separating out the rigid motions. Specifically, we will keep the rigid pose fixed in a canonical form throughout the entire sequence. This means that we only consider variations of the human body due to non-rigid deformations, while rigid movements are not permitted. To accomplish this, it is essential that the network is capable of disregarding shape changes due to rigid motion or changes in viewpoint, and instead, concentrates on predicting shape variations resulting from non-rigid deformations within a consistent canonical view. This is facilitated by employing the concept of the transversal property introduced by \cite{novotny2019c3dpo}. The central idea is to construct a set $\mathcal{X}_0 \in \mathbb{R}^{P \times 3}$ that for any given pair $\mathbf{X}, \mathbf{X}' \in \mathcal{X}_0$, if they are related by a rotation $\mathbf{X'=RX}$, then $\mathbf{X=X'}$.

In single-frame methods like \cite{novotny2019c3dpo} and its follow-up works, they explicitly learn a canonicalization mapping $\Psi: \mathbb{R}^{P \times 3} \rightarrow \mathbb{R}^{P \times 3}$ such that for any arbitrary perturbation $\mathbf{R} \in \text{SO}(3)$ and shape $\mathbf{S} \in \mathbb{R}^{P \times 3}$, it fulfills $\mathbf{S} = \Psi(\mathbf{RS})$. In this way, the network is encouraged to maintain consistency in how reconstructions are executed. However, since their input is single-frame, they have no choice but to aggregate information of all shapes from the dataset. We argue that this way of constructing the transversal set is inappropriate because variations of shapes from separate sequences follow different patterns and should not be grouped together. The transversal set may not suitable for all poses in dataset, while it should only be maintained within the same input sequence. Therefore, to incorporate the transversal property correctly in sequential data, we don’t directly apply $\Psi$ to each individual frame. Instead, we turn to Procrustes analysis, which enables alignment of a set of shapes as closely as possible to a common reference through rigid transformations.
\begin{equation}
\label{eqn:Procrustean alignment}
\begin{aligned}
    &\bar{\mathbf{S}}, \mathbf{R}_i = \mathop{\arg\min}\limits_{\bar{\mathbf{S}}, \mathbf{R}_i} \| \mathbf{R}_i\mathbf{S}_i - \bar{\mathbf{S}} \| \\
    &\text{s.t.} \quad \mathbf{R}_i^\mathrm{T}\mathbf{R}_i  = \mathbf{I}
\end{aligned}
\end{equation}
Here, $\bar{\mathbf{S}}$ represents the reference shape that all other shapes are to be aligned with, and $\mathbf{R}_i$ denotes the corresponding aligning rotation matrices. \cite{park2020procrustean} demonstrated that the set of shapes aligned through Procrustes analysis constitutes a transversal set, which satisfies our design.

Building on this concept, we introduce the Rigid-Motion-Non-Rigid-Deformation (RMNRD) modeling strategy. Given a sequence of shapes $\{\mathbf{S}_1, \mathbf{S}_2,\dots,\mathbf{S}_F\}$, we stabilize their rigid poses by superimposing all shapes onto a common reference shape using Procrustes analysis and construct each original shape $\mathbf{S}_i$ as the sum of the common reference shape $\bar{\mathbf{S}}$ and its corresponding residual deforming shape $\delta_{\mathbf{S}_i}$:
\begin{equation}
\label{eqn:RMNRD basis-deformation}
    \mathbf{S}_i = \bar{\mathbf{S}} + \delta_{\mathbf{S}_i}, i=1,\dots,F
\end{equation}
where $\bar{\mathbf{S}}$ can be computed from Eqn.\ref{eqn:Procrustean alignment}. This modeling strategy not only provides a way of resolving the factorization ambiguity but also bolsters the shape representation capabilities by introducing hierarchical flexibility. Conceptually, the reference shape $\bar{\mathbf{S}}$ is able to capture the low-rank common structure among the sequential 3D shapes, while the residual deforming shape $\delta_{\mathbf{S}_i}$ accommodates diverse minor deformations with enhanced adaptability.

\subsection{Measurement-Conditioned Pose Generation Priors}
\label{diffusion}
To tackle with the inherent ambiguity of the NRSfM pipeline, we propose to utilize a diffusion-based pose generator. Denoising Diffusion Probabilistic Models (DDPM) are a type of generative models that learn the underlying distribution over training data samples. We build our method upon the recent diffusion-based 3D human pose generation model \cite{shan2023diffusion} for its efficiency in generating high-quality human poses with 2D observation guidance. The whole framework consists of a forward diffusion process and a reverse sampling process.

\textbf{Forward diffusion process} perturbs the ground truth 3D pose $y_0$ drawn from an unknown data distribution $p_{data}(y)$ by gradually adding independent Gaussian noise $ \epsilon \sim N(0, I)$ and finally corrupts to a noise pose $y_t$ after $t$ step that follows the predefined prior distribution, e.g. isotropic Gaussian distribution.

\textbf{Reverse diffusion process} samples from the predefined prior distribution and reverse the forward diffusion process to get a plausible pose that satisfies the original data distribution $p_{data}(y)$. The whole diffusion process is trained by minimizing the objective
\begin{equation}
\label{eqn:diffusion optimization}
\begin{aligned}
    \mathbb{E}_{y_0 \sim p_{data}(y),t,\epsilon \sim N(0, I)} [\| y_0 - g_\omega(y_t, c, t) \|_2^2]
\end{aligned}
\end{equation}
where $t$ is a diffusion timestep, $y_0$ is an arbitrary 3D pose drawn from the data distribution and $y_t$ is the noised pose to time t with gaussian noise $\epsilon$. $g_\omega$ is the denoising network with parameters $\omega$. Instead of regressing the noise, we directly regress the original 3D pose. The diffusion model also takes a conditioning input $c$ which serves as a guidance in the reverse process. In our problem setup the condition is the 2D measurements.

Once the diffusion model is well-trained, we'll fix the parameters $\omega$ and use it to instruct the NRSfM optimization process $\mathbf{M}, \mathbf{S} = f_{\theta}(\mathbf{W})$. We can optimize for the predicted shape $\mathbf S$ to follow the pose distribution priors conditioned on $\mathbf W$ and can also backpropagate gradients to the NRSfMFormer and thus get a stochastic gradient descent on $\theta$.

\begin{figure*}[t]
\begin{center}
\includegraphics[width=\linewidth]{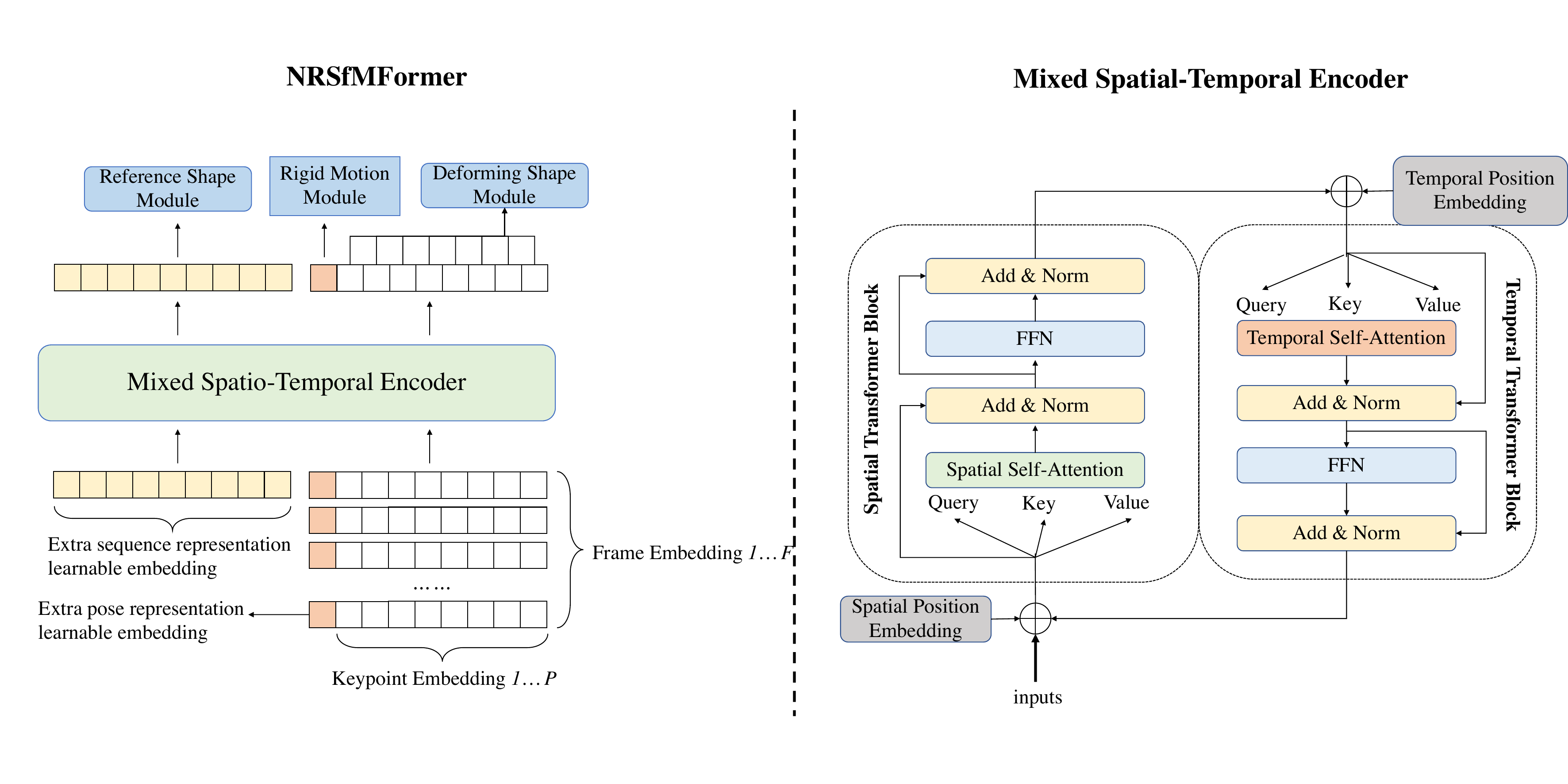}
\caption{Overview of NRSfMFormer. Input is the vector consisting of features from each frame in a video sequence. After being linearly embedded and adding sequence position embeddings, we feed the resulting sequence of vectors to a mixed spatio-temporal transformer encoder. In order to perform reference shape regression, we'll need a sequence representation independent from individual frame features. Therefore, we add an extra learnable sequence representation token to the input. Similarly, we add another learnable embedding to each keypoint feature, which serves as the representation for regressing the object's 6DoF pose. The illustration of the Transformer encoder was inspired by \cite{zhang2022mixste}}
\label{fig:NRSfMFormer}
\end{center}
\vspace{-1em}
\end{figure*}

\subsection{NRSfMFormer}
\label{NRSfMFormer}
We implement the RMNRD framework with NRSfMFormer, a Transformer encoder-based architecture. Its proficiency in modeling dependencies between inputs—regardless of time intervals—makes it highly suitable for leveraging the structure of non-rigid scenes along the temporal dimension. 
The NRSfMFormer is composed of a mixed spatio-temporal encoder for feature extraction and three subnetwork modules that parameterize the components as discussed in previous sections. 

The overview of the NRSfMFormer is depicted in Fig.\ref{fig:NRSfMFormer}.  The network takes the 2D sequence measurement $\textbf{W} \in \mathbb{R}^{F \times P \times 2}$ with $F$ frames and $P$ keypoints as input. We first project the input sequence to a high-dimensional feature $\mathbf{Z} \in \mathbb{R}^{F \times P \times D}$, which can be interpreted at two levels: it consists of $F$ frame features $\mathbf{Z}_i \in \mathbb{R}^{P \times D}, i=1, \dots, F$ and each frame feature consists of $P$ keypoint features $\mathbf{Z}_{ij} \in \mathbb{R}^{D}, j=1, \dots, P$. Then we apply a position embedding for both spatial and temporal domains to retain the positional information. 

Different from previous work using the Transformer structure, NRSfMFormer adopts a novel way of generating a unified feature representation that incorporates information required by both human skeleton estimation and 6DoF rigid transformation estimation tasks. Specifically, NRSfMFormer considers reference shape as one of the properties of the pose sequence, and similarly, the rigid transformation between each camera position and the human skeleton is a property that is bound to the pose in each individual frame. Enlightened by ViT’s design of $[class]$ token \cite{dosovitskiy2020image}, we attach two learnable embeddings to both frame-level and keypoint-level features so that the final input to the encoder is:
\begin{equation}
\begin{aligned}
\label{eqn:transformer input}
    \textbf{Z} &= [\mathbf{Z}_0, \mathbf{Z}_1, \mathbf{Z}_2, \dots, \mathbf{Z}_F] \in \mathbb{R}^{(F + 1) \times (P + 1) \times D}
\end{aligned}
\end{equation}
The design intuition behind this is that the embedded token added on the frame feature serves as the semantic representation for the whole sequence $(\mathbf{Z}_0 = x_{\text{seq}})$. Similarly, the embedded token added on the keypoint feature serves as the representation for regressing the object's 6DoF rigid transformation $(\mathbf{Z}_{i0} = x_{\text{pose}})$.

Following the design scheme from \cite{zhang2022mixste}, we stack $L$ Mixed Spatio-Temporal Encoder block coherence feature extraction. It alternately uses Spatial Transformer Block (STB) and Temporal Transformer Block (TTB) to learn spatial and temporal correlation with multi-head self-attention. STB computes the attention between joints and aims to learn the body joint relations within each frame. On the other hand, TTB computes the attention between frames and focuses on learning the global temporal correlation of each joint across frames.

Passing through the encoder blocks, we now have feature $\textbf{Z}^L = [Z_0^L, Z_1^L, Z_2^L, \dots, Z_F^L]$ where the superscript indicates which encoder layer we're extracting the feature from. On top of that, subsequent MLP-based modules \textbf{Rigid Motion Module}, \textbf{Deforming Shape Module} and \textbf{Reference Shape Module} are used to recover the geometric meaning of corresponding features from the latent space. Specifically, the rigid motion module and the reference shape module are applied to the extra learnable embedded tokens $Z_{i0}^L$ and $Z_{0}^L$ to regress 6DoF rigid transformation and reference shape. The deforming shape module is applied to the remaining individual representations from Transformer encoder output $\{Z_{i1}^L, Z_{i2}^L, \dots, Z_{iP}^L\}, i=1, \dots, F$ to predict the residual shape corresponding to each 2D input measure with respect to the reference shape.
\begin{equation}
\label{eqn:regression head}
\begin{aligned}
    \bar{\mathbf{S}} &= \text{MLP}(Z_0^L)) \\
    \delta_{\mathbf{S}_i} &= \text{MLP}(Z_{i}^L)) \quad\quad i=1,\dots,F \\
    \mathbf{R}_i &= \text{MLP}(Z_{i0}^L)) \quad\quad i=1,\dots,F
\end{aligned}
\end{equation}

\subsection{Loss Design}
\label{Loss}
To solve the NRSfM problem defined in Sec.\ref{problem formulation}, we propose to minimize a differentiable loss function with respect to motion parameters $\textbf{R} = \{\mathbf{R}_1,\mathbf{R}_2,\dots,\mathbf{R}_F\}$ and human pose $\textbf{S}=\{\mathbf{S}_1,\mathbf{S}_2,\dots,\mathbf{S}_F\}$. In the following section, we'll introduce each part separately.

\noindent\textbf{Reprojection loss} Reprojection loss is utilized to enforce that the network learns accurate motion and shape estimations in a self-supervised manner, ensuring alignment with 2D observations.
\begin{equation}
    \mathcal{L}_{reproj} = \frac{1}{F} \sum_{i=1}^F \| \mathbf{\Pi R_iS_i - W_i} \|_F^2
\end{equation}

\noindent\textbf{Procrustean alignment loss} According to the analysis in Sec.\ref{disentagle}, one crucial ingredient of the RMNRD modeling strategy is that all shapes in the sequence should align to a common reference shape through Procrustes analysis so that the transversal property can be retained. 

Specifically, with our predicted shapes, reference shapes, and residual shapes $\mathbf{S}_i^{pred}, \bar{\mathbf{S}}^{pred}, \delta_{\mathbf{S}_i}^{pred}$, we can follow the procedure described in algorithm \ref{alg:gpa_solver} for solving a Procrustean alignment problem to associate $\bar{\mathbf{S}}^{pred}$ and $\mathbf{S}_i^{pred}$ as close as possible. This enables us to guarantee that by imposing constraints based on Procrustean alignment, for a sequence of shapes whose variations only exist in non-rigid deformations, we can always eliminate the ambiguity caused by rigid motions.
\begin{equation}
    \mathcal{L}_{proc} = \frac{1}{F} \sum_{i=1}^F \| \mathbf{S}_i^{pred} - \mathbf{S}_i^{*} \|
\end{equation}
where $S_i^{*}$ is the shape retrieved after Procrustean alignment.

\begin{algorithm}[ht!]
\caption{Procrustean Alignment} 
\label{alg:gpa_solver}
\begin{algorithmic}
\renewcommand{\algorithmicrequire}{\textbf{Input:}}
\renewcommand{\algorithmicensure}{\textbf{Output:}}

\REQUIRE Shape sequence $\{\mathbf{S}_i\}$, reference shape $\bar{\mathbf{S}}$
\ENSURE Aligned shape sequence $\{\mathbf{S}_i^*\}$

\STATE Centralize each shape as $\mathbf{S}_i \leftarrow \mathbf{S}_i - \mathbf{S}_i \textbf{1}\textbf{1}^T$
\STATE Centralize reference shape as $\bar{\mathbf{S}} \leftarrow \bar{\mathbf{S}} - \mathbf{S}_i \textbf{1}\textbf{1}^T$

\STATE Update the rotations that can align each shape to the reference shape $\mathbf{R}_i = \mathbf{U}_i\mathbf{V}_i^T$, where $\bar{\mathbf{S}}^TS_i = \mathbf{U}_i \mathbf{\Lambda} \mathbf{V}_i^T$  is obtained through singular value decomposition.
\STATE Update aligned shapes as $\mathbf{S}_i^* \leftarrow \mathbf{R}_i\mathbf{S}_i$

\end{algorithmic}
\end{algorithm}

\noindent\textbf{Diffusion prior loss} As discussed in Sec.\ref{diffusion}, the output of NRSfMFormer should satisfy certain prior distribution defined by pretrained diffusion model $g_\omega$.
\begin{equation}
\begin{aligned}
    \mathcal{L}_{prior} &= \mathbb{E}_{t,\epsilon \sim N(0, I)} [\| y_0 - g_\omega(y_t, \mathbf{W}, t) \|_2^2] \\
    y_0 &= f_\theta(\mathbf{W})
\end{aligned}
\end{equation}
This term is similar to Eqn.\ref{eqn:diffusion optimization}, despite that $y_0$ is not sampled arbitrarily from the data distribution but is the output of our NRSfMFormer. In addition, we use the 2D measurement matrix $\mathbf W$ as the condition during the diffusion process.

\noindent\textbf{Temporal smoothness loss} As we're using sequence data as input, we propose to consider the temporal smoothness assumption as the object moves and deforms continuously in the physical world.
\begin{equation}
    \mathcal{L}_{smooth} = \sum_{i=2}^F \lambda_\mathbf{R} \| \mathbf{R}_i - \mathbf{R}_{i-1} \|^2 + \lambda_\mathbf{S} \| \mathbf{S}_i - \mathbf{S}_{i-1} \|^2
\end{equation}



Combining all loss functions introduced above, our total loss is as follows:
\begin{equation}
\label{eqn:loss function}
    \mathcal{L} = \beta_1 \mathcal{L}_{reproj} + \beta_2 \mathcal{L}_{proc} + \beta_3 \mathcal{L}_{prior} + \beta_4 \mathcal{L}_{smooth}
\end{equation}
where $\beta_1, \beta_2, \beta_3, \beta_4$ are trade-off weights hyperparameters.

\section{Experiments}
In this section, we'll introduce our experimental results evaluated on different large-scale human pose estimation datasets. We first introduce the datasets and evaluation metrics we are using and then the implementation details. Next, qualitative experimental results and comparisons with other methods are reported. Finally, we analyze the influence of different components of the proposed method in detail through ablation studies.

\subsection{Datasets and Evaluation Protocols}
\noindent\textbf{Human3.6M}\cite{ionescu2013human3} is a large dataset containing various types of human motion sequences annotated with 3D ground truth extracted using motion capture systems. The 3D human skeleton in Human3.6M is modeled as a 17-joint skeleton, subjects S1, S5, S6, S7, and S8 are applied during training, while the S9 and S11 are used for testing. 

\noindent\textbf{MPI-INF-3DHP}\cite{mehta2017monocular} is a recently popular dataset that features in more complex scenarios including indoor, green screen, and outdoor scenes. The training set contains 8 activities performed by 8 actors, and the test set covers 7 activities. Following \cite{shan2022p}, we use the valid frames provided by the official for testing. 

\noindent\textbf{Evaluation metrics} We report the mean per joint position error (MPJPE) in millimeters. Since an unsupervised setting does not contain metric data we scale the reconstructed 3D pose to match ground truth, commonly known as N-MPJPE. Besides, Procrustes alignment (including scaling), known as PA-MPJPE is also reported. For 3DHP datasets, we also report the Percentage of Correct Keypoints (PCK) thresholded at 150mm and the Area Under the Curve (AUC).

\subsection{Implementation Details}
Our model is implemented in a two-stage training scheme. In the first stage, we train the DDPM model for 40 epochs with AdamW optimizer and a learning rate of $5e^{-5}$. In the second stage, we freeze the diffusion model and only train the NRSfMFormer part. We adopt a similar architecture as that of Zhang et al. \cite{zhang2022mixste}, which consists of 4 mixed spatio-temporal encoder blocks. All the regression heads are single-layer MLP. All parameters are initialized randomly from the zero-mean Gaussian distribution with $\sigma = 0.001$. During training, the input sequence length is 27 and we use AdamW optimizer with a mini-batch size of 128 to update the parameters. We trained 50 epochs in total with $5e^{-5}$ initial learning rate, and the learning rate will reduce by a factor of 0.99 after each epoch. Our experiments are performed on one NVIDIA 1080Ti GPU, and the training takes about 20 hours to complete.

\subsection{Quantitative \& Qualitative results}

\begin{table}[t]
\centering
\small
\setlength{\tabcolsep}{0.32cm}
\begin{tabular}{ l | c | c }
\toprule
 \textbf{Unsupervised 3D HPE} & N-MPJPE $\downarrow$ & PA-MPJPE $\downarrow$\\
\midrule
  Chen \etal\cite{chen2019unsupervised} & - & 51\\
  Yu \etal\cite{yu2021towards} & 85.3 & 42.0 \\
  Wang \etal\cite{wandt2022elepose} & 64.0 & 36.7 \\
  \textbf{NRSfMFormer} & \underline{41.6} & \underline{33.1}\\
\midrule
\midrule
 \textbf{NRSfM} & N-MPJPE $\downarrow$ & PA-MPJPE $\downarrow$\\
\midrule
    Park \etal\cite{park2020procrustean} & 86.4 & - \\
    Wang \etal\cite{wang2021paul} & 88.3 & - \\
    Novotny \etal\cite{novotny2019c3dpo} & 95.6 & - \\
    Xu \etal\cite{xu2021invariant} & 77.2 & - \\
    Deng \etal\cite{deng2022deep} & 79.8 & - \\
    Zeng \etal\cite{zeng2022mhr} & 72.6 & - \\
    \textbf{NRSfMFormer} & \underline{41.6} & \underline{33.1}\\
\bottomrule
\end{tabular}
\caption{3D HPE quantitative results on Human3.6M dataset. All results are reported in millimeters (mm). The diffusion prior model is trained on Human3.6M and GT 2D keypoints inputs are used.}
\label{tab:h36m}
\end{table}

\begin{figure*}[t]
\begin{center}
\includegraphics[width=\linewidth]{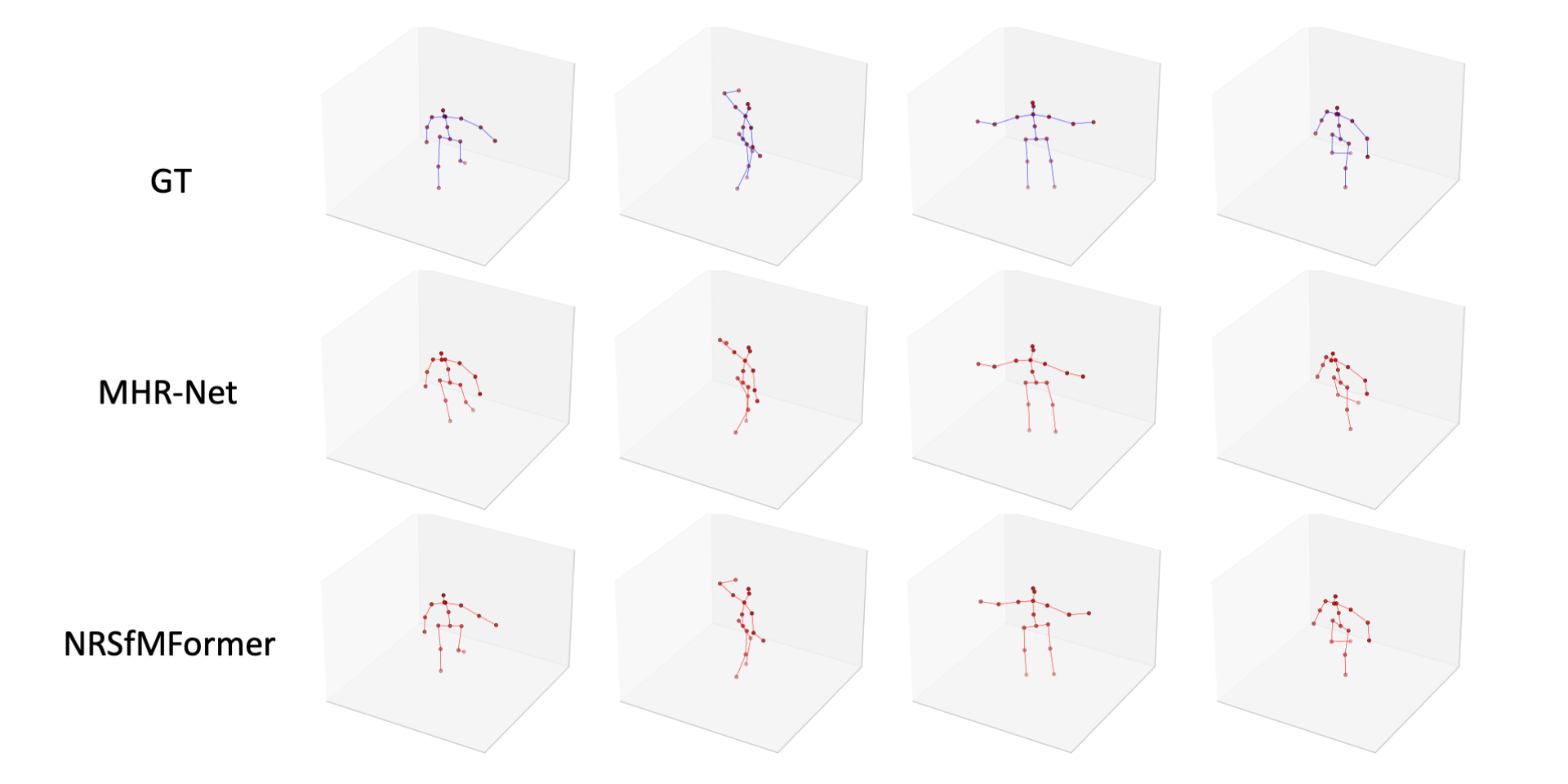}
\caption{Visualization of several methods on Human3.6M with ground truth 2D keypoint as input.}
\label{fig:Visualization}
\end{center}
\end{figure*}

\noindent\textbf{Results on Human3.6M} We compare our model with unsupervised 3D HPE methods as well as NRSfM methods that are evaluated on the Human3.6M dataset. All experiments use the GT 2D keypoints as input and our diffusion prior is pretrained on the training set of Human3.6M. As shown in Table \ref{tab:h36m}, we obtain $41.6.4$mm in N-MPJPE and $33.1$mm in PA-MPJPE, which outperforms any existing SOTA methods in both unsupervised 3D HPE and NRSfM by a large margin. Be advised that most of the comparing methods use single-frame data as input to predict the corresponding 3D pose. In contrast, our method explicitly models the underlying sequence structure with the temporal constraint, taking advantage of the information ignored by the other methods and thus achieving improved results. We also show some qualitative results in Fig.\ref{fig:Visualization} comparing with the SOTA NRSfM method \cite{zeng2022mhr}. From the figure we can observe that our model has more accurate predictions especially on edge joints of the human body, like hands, feet, etc. This contributes to the performance improvements of our model.

\begin{table}[t]
\centering
\small
\setlength{\tabcolsep}{0.23cm}
\begin{tabular}{ l | c | c | c}
\toprule
 \textbf{Unsupervised 3D HPE} & PA-MPJPE $\downarrow$ & PCK $\uparrow$ & AUC $\uparrow$\\
\midrule
  Chen \etal\cite{chen2019unsupervised} & - & 71.7 & 36.3\\
  Yu \etal\cite{yu2021towards} & - & 86.2 & 51.7 \\
  Wang \etal\cite{wandt2022elepose} & 54.0 & \underline{86.0} & 50.1\\
  \textbf{NRSfMFormer} &\underline{46.3} & 73.4 & \underline{68.1}\\
\midrule
\end{tabular}
\caption{Our results (14-joint) on MPI-INF-3DHP with metrics as in Wang \etal\cite{wandt2022elepose}. All results are reported in millimeters (mm). GT 2D keypoints inputs are used.}
\label{tab:3dhp}
\vspace{-0.75em}
\end{table}

\noindent \textbf{Results on 3DHP}. Following previous works\cite{chen2019unsupervised, yu2021towards, wandt2022elepose}, we use ground truth 2D poses as input. As shown in Table\ref{tab:3dhp}, we achieve $46.3$mm in PA-MPJPE, which outperforms existing SOTA performance by $14\%$. Although there exists some gap in terms of PCK, our method prevails in AUC by a margin up to nearly $40\%$, which still demonstrates the effectiveness of our method.

\subsection{Ablation Study}


In this section, we study the impact of different building blocks in our model. All results are evaluated on Human3.6M dataset and with the same configuration and training scheme.

We set up several ablated models to study the effect of RMNRD modeling strategy, temporal consistency, procrustean alignment, and diffusion prior. As shown in Table \ref{tab:ablation}, we observe the degradation of performance by removing any of the components, indicating the effectiveness of our general framework design. Among all the compositions, we can see that the most crucial one is the diffusion model. Without the pretrained prior, performance will drop dramatically and errors will even double. This reveals the fact that resolving the depth ambiguity and predicting a plausible 3D pose should be the main concern when solving an NRSfM problem. Other constraints like motion ambiguities and temporal information will also have respective contributions but the impacts are not as significant.

\section{Conclusion}
In this paper, we aim to solve the sequential unsupervised 3D human pose estimation problem with Non-Rigid Structure-from-Motion modeling. Inspired by the field of non-rigid structure-from-motion, we design a geometric-aware pipeline that uses a new modeling scheme for human pose deformations, which divides the task of 3D HPE into the estimation of the 3D sequence reference skeleton and a frame-by-frame skeleton deformation. We also propose to use a diffusion-based shape prior to resolve the inherent depth ambiguity. One limitation of this work is that NRSfM is essentially a universal method that can be applied to diverse scenarios. However, in this work we're only focusing on its application to the human body and not fully utilizing the potential of either NRSfM or diffusion prior. In future  work and will expand our test subjects to more general cases.

\begin{table}[t]
\centering
\small
\tabcolsep=0.32cm
\begin{tabular}{lcc}
\hline
Ablations      & N-MPJPE $\downarrow$ & PA-MPJPE $\downarrow$ \\ \hline
w/o RMNRD strategy     & 46.7       & 36.5   \\
w/o temporal consistency     & 42.3     & 33.8   \\ 
w/o Procrustean alignment       & 44.6       & 36.7   \\
w/o diffusion prior           & 77.3          & 54.4   \\
\textbf{NRSfMFormer}     & \underline{41.6} & \underline{33.1}  \\ 
\hline
\end{tabular}

\caption{Ablation studies performed on the Human3.6M dataset}
\label{tab:ablation}
\vspace{-1.5em}
\end{table}

{\small
\bibliographystyle{ieee_fullname}
\bibliography{egbib}
}

\end{document}